%% file: main.tex
\documentclass[10pt,twocolumn,letterpaper]{article}

\usepackage{cvpr2016pkg/cvpr}

\usepackage{times}
\usepackage{epsfig}
\usepackage{graphicx}
\usepackage{amsmath}
\usepackage{amssymb}
\usepackage{booktabs}
\usepackage{latex_pkgs/msmaths}
\usepackage{latex_pkgs/msformatting}
\usepackage[conf]{optional}

\usepackage{caption}
\usepackage{cuted}

\usepackage[pagebackref=true,breaklinks=true,letterpaper=true,colorlinks,bookmarks=false]{hyperref}

\cvprfinalcopy 

\def\cvprPaperID{2441} 

\begin{document}

\title{
Spatio-temporal Human Action Localisation and\\
Instance Segmentation in Temporally Untrimmed Videos
}


\author
{
Suman Saha$^1$\\
\and
Gurkirt Singh$^1$\\
\and
Michael Sapienza$^2$\\
\and
Philip H. S. Torr$^2$\\
\and
Fabio Cuzzolin$^1$\\
$^1$Oxford Brookes University \quad $^2$University of Oxford\\
{\tt\small \{suman.saha-2014, gurkirt.singh-2015, fabio.cuzzolin\}@brookes.ac.uk}\\  
{\tt\small \{michael.sapienza, philip.torr\}@eng.ox.ac.uk}\\
}
\maketitle
\begin{strip}
  \centering
  \includegraphics[width=0.95\textwidth]{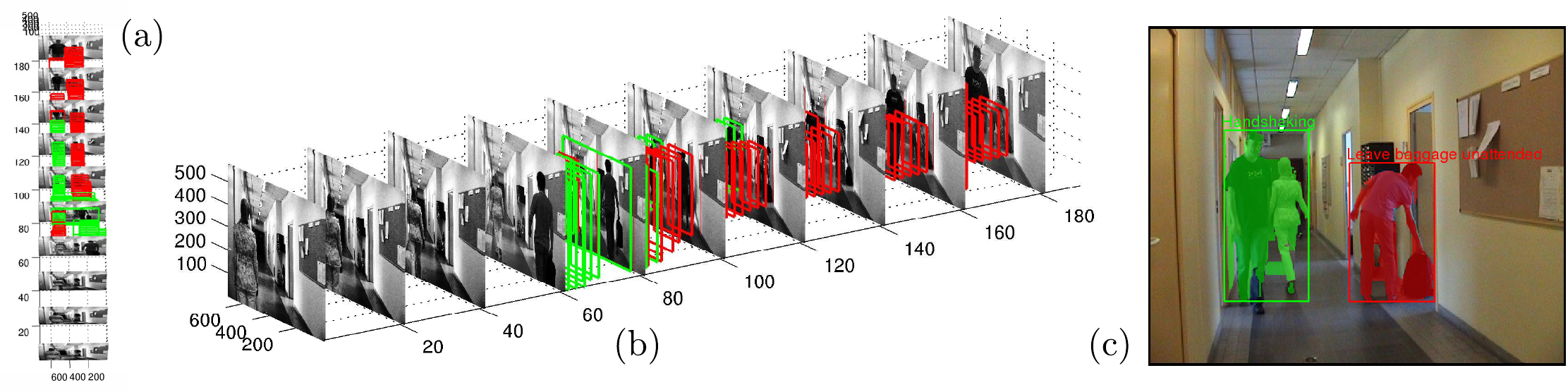}
  \vskip 0.1cm
  \begin{minipage}[adjusting]{0.95\textwidth}
    Figure 1: 
    A video sequence taken from the LIRIS-HARL dataset plotted in space-and time.
    \textbf{(a)} A top down view of the video plotted with the detected action tubes of class `handshaking' in green, and `person leaves baggage unattended' in red. Each action is located to be within a space-time tube.
    \textbf{(b)} A side view of the same space-time detections. Note that no action is detected at the beginning of the video when there is human motion present in the video.
    \textbf{(c)} The detection and instance segmentation result of two actions occurring simultaneously in a single frame.
  \end{minipage}
  \label{fig:introductionTeaser}
\end{strip}
\setcounter{figure}{1}

\begin{abstract}  
\vskip 0.3cm
  \input{text/abstract}

\end{abstract}
\input{text/intro}

\input{text/contrib}

\input{text/soa}
\input{text/approach}
\input{text/datasets_performance_measures}
\input{text/experiments_results_discussion}
\input{text/conclusion}

{\small
\bibliographystyle{cvpr2016pkg/ieee}
\bibliography{refcvpr2016}
}

\end{document}

%% file: text/abstract.tex
\vskip -1.2cm
Current state-of-the-art human action recognition is focused on the classification of temporally trimmed videos in which only one action occurs per frame.
In this work we address the problem of action localisation and instance segmentation in which multiple concurrent actions of the same class may be segmented out of an image sequence.
We cast the action tube extraction as an energy maximisation problem in which configurations of region proposals in each frame are assigned a cost and the best action tubes are selected via two passes of dynamic programming.
One pass associates region proposals in space and time for each action category, and another pass is used to solve for the tube's temporal extent and to enforce a smooth label sequence through the video.
In addition, by taking advantage of recent work on action foreground-background segmentation, we are able to associate each tube with class-specific segmentations.
We demonstrate the performance of our algorithm on the challenging LIRIS-HARL dataset and achieve a new state-of-the-art result which is 14.3 times better than previous methods.

%% file: text/intro.tex
\section{Introduction}
Emerging real-world applications require an all-round approach to the machine understanding of human behaviour which goes beyond the recognition of simple, isolated actions from video.
Existing works on action recognition have achieved impressive recognition rates, however they are mostly focused on action classification~\cite{laptev-2008,wang-2011,wang-2013, Shuiwang-2013,Karpathy-2014,Simonyan-2014} and localisation~\cite{Georgia-2015a, Weinzaepfel-2015} in settings where each video clip contains only a single action category and in which videos are temporally trimmed by human observers.
In contrast, here we consider real-world scenarios where videos often contain concurrent instances of multiple actions or no action at all.
This makes the problem significantly harder, as one needs to concurrently solve the following problems:
i) generate 3D tube proposals to bound the location of a possible action, and
ii) classify each space-time tube candidate into one of several action categories.
An as example, consider the example shown in Fig.~1,
where we detect multiple actions (``leaving bag unattended'' and ``handshaking'') occurring simultaneously in different space-time locations.
The video is taken from the LIRIS-HARL action detection dataset \cite{liris-harl-2012} which poses several additional challenges since many actions
 i) have similar appearance and motion whilst belonging to distinct classes such as `unsuccessfully unlocking door' and 'successfully unlocking door', and 
ii) have very different appearance yet share the same class such as `put/take object into/from box'.

%% file: text/contrib.tex
\paragraph{Contributions}
In this paper we propose an algorithm for human action detection and segmentation in which configurations of region proposals in each frame are assigned a cost and the best action tubes are selected via two passes of dynamic programming.
Moreover, by taking advantage of the human foreground-background segmentation work by \cite{Lu_2015_CVPR},
we generate action frame proposals based on the power set of connected components in the foreground-background segmentation.
This means that we can output a pixel-level action instance segmentation in addition to detection with tubes.
To the best of our knowledge our algorithm provides the best human action detection results on the most challenging dataset available to date.
Lastly, we are the first to show qualitative action instance segmentation results.



%% file: text/soa.tex
\section{Related work}

Many approaches to action recognition \cite{laptev-2008,wang-2011,wang-2013} are based on appearance (e.g., HOG \cite{dalal-2005} or SIFT \cite{Lowe-2004}) or motion features (e.g., optical flow, MBH \cite{dalal-2006}),
encoded using Bag of Visual Words or Fisher vectors.
The resulting descriptors are typically used to train classifiers (e.g. SVM) in order to predict the labels of action videos.
Recently, however, inspired by the record-breaking performance of CNNs in image classification \cite{krizhevsky2012} and object detection from images \cite{girshick-2014}, deep learning architectures have been increasingly applied to action classification \cite{Shuiwang-2013,Karpathy-2014,Simonyan-2014} and localisation \cite{Georgia-2015a,Weinzaepfel-2015}.

For instance in global action classification,
Simonyan and Zisserman \cite{Simonyan-2014} have proposed a novel feature extraction approach based on two Convolutional Neural Networks (CNNs),
one encoding static appearance features from RGB images, and the other extracting motion features from optical flow heat maps.
Gkioxari and Malik \cite{Georgia-2015a} have extended the work of \cite{girshick-2014} and \cite{Simonyan-2014} to tackle both classification and localisation.
Moreover, Weinzaepfel at al. \cite{Weinzaepfel-2015} use a tracking-by-detection approach based on a novel track-level descriptor (Spatio-Temporal Motion Histogram, STMH) combined with CNN features.
The downside of the aforementioned approaches is that they perform localisation of actions in videos which only contain one action.
Furthermore, the action videos are already temporally trimmed and therefore it is hard to evaluate their temporal localisation.

In order to improve action recognition and localisation, Georgia  \etal\  \cite{Georgia-2015b}  and and Jain  \etal\  \cite{Jain-2015},
use CNN features and add contextual cues.
Whereas the work of Georgia \etal\ is limited to localisation in still images,
the approach by Jian \etal\ will only reap benefits where the objects in the action are discriminative.
In some of the actions in the LIRIS-HARL dataset, categories such as `unsuccessfully unlocking door' and `successfully unlocking door' both have the same set of objects. In this case, the object `door' does not reveal any additional information about the action.



The temporal localisation actions~\cite{jiang2014thumos,gorban2015thumos}, events~\cite{TRECVID} and gestures~\cite{escalera2014chalearn} in temporally untrimmed videos has attracted much attention recently.
These challenges led to big advances in the state-of-the-art \cite{yeung2015every,xu2014discriminative}.
However, unlike our work, these approaches address only the temporal action localisation.
For multiple co-occurring actions in time, Yeung \etal\ introduced a Long Short Term Memory network \cite{yeung2015every}.
They augmented the annotation on the Thumos Dataset \cite{jiang2014thumos} to include new categories and co-occurring actions;
the downside is that spatial localisation information has been ignored.

Multiple concurrent action detection from temporally untrimmed videos has only been explored on a small number of action classes.
For instance Laptev \etal\ proposed an action detection approach based on \textit{keyframe priming}~\cite{laptev-2007}.
Their idea was to improve space-time interest point detectors for actions such as `drinking',
with single frame detection from state-of-the-art methods in object detection.
Alternatively, \cite{klaser-2010} split action detection into two parts: i) detecting and tracking humans,
and ii) using a space-time descriptor and sliding window classifier to detect the location of two actions (phoning and standing up).
By contrast we consider 10 different action categories.

In this paper we propose a novel action detection method which addresses the challenges involved in classification, localisation and detection of co-occurring actions in space-and-time given temporally untrimmed videos. 


%% file: text/approach.tex
\section{Methodology} \label{sec:methodology}

\begin{figure*}[t]
  \centering
  \includegraphics[width=\textwidth]{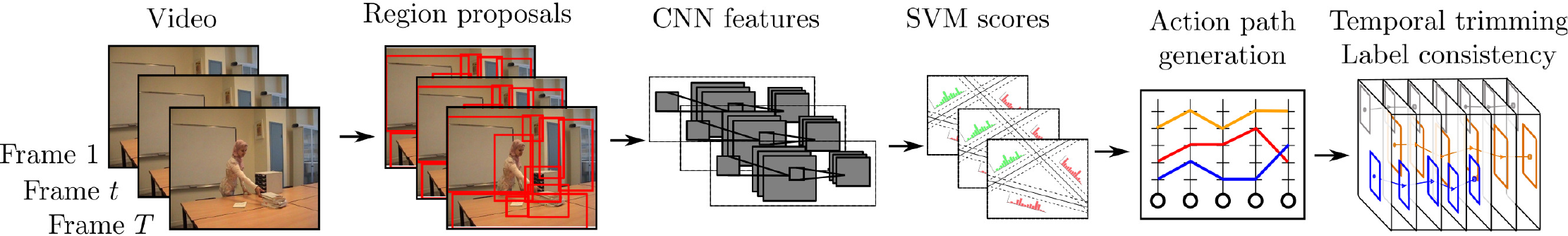}
  \caption{Overview of the proposed action detection pipeline. }
  \label{fig:algorithmOverview}
\end{figure*}

The following methodology takes as input raw image frames and generates action-specific space-time tubes.
An overview of the algorithm is depicted in Fig.~\ref{fig:algorithmOverview}.
\subsection{Problem formulation} \label{subsec:prob_formulation}
We start by extracting region proposals per frame (\S~\ref{sec:regionProposals}), 
followed by CNN feature extraction (\S~\ref{sec:featureExtraction}).
The features extracted from each region proposal are subsequently scored by a support vector machine \cite{girshick-2014} (\S~\ref{sec:regionScoring}).
In order to associate the region proposals in space and time, 
two passes of dynamic programming are used (\S~\ref{sec:dynamicProgramming}).
The final result is a set of space-time tubes to denote the detection,
and associated pixel sets which denote the 3D instance segmentations.
We define an action tube $\Set{T}$ as a sequence of contiguous region proposals connected over time $t$ without any holes.

We define a video $\Set{V}$ as a sequence of frames $(f_{\Index{1}}, \dots, f_{T})$~(cf.~Fig.~\ref{fig:algorithmOverview}).
The task is to detect multiple concurrent actions, and thus, we start by identifying regions in each image frame which are highly likely to contain human actions.
We denote each 2D region proposal (\S~\ref{sec:regionProposals}) `$\Vector{r}$' as a subset of the image pixels,
associated with a minimum bounding box `$\Vector{b}$' around it.
We further assume that each region proposal has a set of class-specific scores $\Vector{s}_{\Scalar{c}}$,
where $c$ denotes the action category label, $\Scalar{c} \in \{1, \dots, \Constant{C}\}$.
Given a set of 2D region proposals in an entire video, we would like to identify sequences of regions most likely to form action tubes.

We cast the action tube extraction as an energy maximisation problem in which configurations of region proposals in each frame are assigned a cost and the best action tubes are selected via two passes of dynamic programming.
In the first pass, we construct paths $\Vector{R}_{c}$ by associating the region proposals over time using their class-specific scores and their temporal overlap as unary potentials.
Candidate action paths $\Vector{R}_{c}=\{ \Vector{r}_{\Index{1}}, \dots,  \Vector{r}_{T} \}$ initially form a sequence of consecutive region proposals spanning the entire length $T$ of the video.
We use a second pass of DP to localise each action in time and to ensure the paths are relatively smooth and have consistent labellings.
The final detection results are found by selecting those tubes with the greatest scores.

\subsection{Region proposal generation} \label{sec:regionProposals}
We use two competing approaches to generate region proposals for action detection.
The first is based upon Selective Search \cite{uijlings-2013}, and the second is adapted from the human motion segmentation work by \cite{Lu_2015_CVPR}.
Whilst using the Selective Search based method for both training and testing,
we only use the motion segmentation based method for testing since it does not provide good negative proposals to use during training. 
Having a sufficient number of negative examples is crucial to train an effective classifier.
At test time, Lu et al.~\cite{Lu_2015_CVPR}'s method allows us to extract pixel-level {action instance segmentation}' information which is superior to what we may obtain by using Selective Search.
We validate our action detection pipeline using both algorithms - the results are discussed in Section~\ref{subsec:discussion}.

\paragraph{Measuring ``actionness'' of Selective Search proposals.} \label{sec:actionness}
The selective-search region-merging similarity score is based on a combination of colour (histogram intersection), and size properties, encouraging smaller regions to merge early, and avoid holes in the hierarchical grouping.

Selective Search (SS) generates on average 2,000 region proposals per frame, most of which do not contain human activities.
In order to rank the proposals with an actionness score and prune irrelevant regions,
we compute dense optical flow between each pair of consecutive frames using the state-of-the-art algorithm in~\cite{Brox-2004}.
Unlike Gkioxari and Malik~\cite{Georgia-2015a}, we use a relatively smaller motion threshold value to prune {SS} boxes,
to avoid neglecting human activities which exhibit minor body movements exhibited in the LIRIS HARL~\cite{liris-harl-2012} such as ``typing on keyboard'', ``telephone conversation'' and ``discussion'' activities.
In addition to pruning region proposals, the 3-channel optical flow values (i.e., flow-$x$, flow-$y$ and the flow magnitude) are used to construct `motion images' from which CNN motion features are extracted~\cite{Georgia-2015a}.

\paragraph{Human motion segmentation (HMS) proposals.} \label{subsec:human_mot_seg}
The human motion segmentation \cite{Lu_2015_CVPR} algorithm  generates binary segmentation of human actions.
It extracts human motion from video using long term trajectories~\cite{Brox_2011}.
In order to detect static human body parts which don't carry any motion but are still significant in the context of the whole action,
it attaches scores to these regions using a human shape prior from a deformable part-based (DPM) model~\cite{felzenszwalb-2010}. 
By striking balance between the human motion and static human-body appearance information,
it generates binary silhouettes of human actions in space and time.
At test time our region proposal algorithm accepts the binary segmented images produced by ~\cite{Lu_2015_CVPR},
and generates region proposal hypotheses $\Vector{r}_{\Index{i}}$ using all possible combinations of 2D connected components ($2^N - 1$) present in the binary map.

%
\subsection{Appearance and motion CNN descriptors}  \label{sec:cnnDescriptor}
In the second stage of the pipeline, we use the ``actionness'' ranked region proposals (cf.~\ref{sec:actionness}) to select image patches from both the RGB (original video frames) and flow images.
The image patches are then fed to a pair of fine-tuned Convolutional Neural Networks (which encode appearance and local image motion, respectively) from which appearance and flow feature vectors were extracted.
As a result the first network learns static appearance information (both lower-level features such as boundary lines, corners, edges and high level features such as object shapes), while the other encodes action dynamics at frame level.
The output of the Convolutional Neural Network may be seen as a highly nonlinear transformation $\Phi(.)$ from local image patches to a high-dimensional vector space in which discrimination may be performed accurately even by a linear classifier.
We follow the network architectures of~\cite{krizhevsky2012} and~\cite{Zeiler2013}. 

\textbf{Pre-training.}
We adopt a CNN training strategy similar to \cite{girshick-2014}.
Indeed, for domain-specific tasks on relatively small scale datasets, such as LIRIS HARL~\cite{liris-harl-2012}, it is important to initialise the CNN weights using a model \emph{pre-trained on a larger-scale dataset},
in order to avoid over-fitting~\cite{Georgia-2015a}.
Therefore, to encode object ``context'' we initialise the appearance-based CNN's weights using a model pre-trained on the PASCAL VOC 2012's object detection dataset.
To encode typical motion patterns over a temporal window, the optical flow-based CNN is initialised using a model pre-trained on the UCF101 dataset (split 1)~\cite{soomro-2012}.\\
\textbf{Fine tuning.}
To fine-tune the pre-trained domain-specific appearance- and flow-based CNNs using LIRIS HARL's training date we use Caffe~\cite{Jia-2013}.
There is a total of 23,910 training frames in the LIRIS HARL dataset.
For training CNNs, the region proposals with an intersection-union overlap score greater than 0.5 with respect to the ground truth bounding box were considered as positive examples, the rest as negative examples.
The image patches specified by the minimum bounding box around the pruned region proposals were randomly cropped and horizontally flipped by the Caffe's \emph{WindowDataLayer}~\cite{Jia-2013} with a crop dimension of $227\times227$ and a flip probability of $0.5$.
Random cropping and flipping was done for both RGB and flow images~(cf. \ref{sec:actionness}).
The pre-processed image patches were then passed to the related CNNs to extract feature vectors from them.
A mini batch of 128 image patches (32 positive and 96 negative examples) are processed by the CNN at each time.
Note that the number of batches varies frame-to-fame as per the number of ranked proposals per frame.
It makes sense to include fewer positive examples (action regions) as these are relatively rare when compared to background patches (negative examples).\\
\textbf{Feature extraction from CNN layers.} 
\label{sec:featureExtraction}
We extract the appearance- and flow-based features from the \emph{fc7} (fully connected layer 7) layer of the the two networks.
Thus, we get two feature vectors (each of dimension 4096): appearance feature `$\Vector{x}_{a}=\Phi_{a}(\Vector{r})$' and flow feature `$\Vector{x}_{f}=\Phi_{f}(\Vector{r})$'.\\
We perform L2 normalisation on the obtained feature vectors, to then, scale and merge appearance and flow features in an approach similar to that proposed by~\cite{Georgia-2015a}.
This yields a single feature vector $\Vector{x}$ for each image patch $\Vector{r}$.
Such frame-level region feature vectors are used to train an SVM classifier (Section \ref{sec:regionScoring}).
\subsection{Training region proposal classifiers} \label{sec:regionScoring}
Once discriminative feature vectors $\Vector{x} \in \mathbb{R}^n$ are extracted from region proposals (cf.~\ref{sec:regionProposals}), 
they can be used to train a set of binary classifiers to attach a vector of scores $\Vector{s}_{c}$ to each region proposal `$\Vector{r}$', where each element in the score vector $\Vector{s}_{c}$ is a confidence measure of each action class $\Scalar{c} \in \{1, 2, \dots, \Constant{C}\}$ to be present within that region.
Due to the recent success of linear SVM classifiers when combined with CNN features \cite{girshick-2014}, we trained a set of 1-vs-rest linear SVMs to classify region proposals.

\textbf{Class specific positive and negative examples.}
In contrast to the RCNN-based one-vs-rest training approach of \cite{girshick-2014},
in which only the the ground-truth bounding boxes are considered as positive examples, due to extremely high inter- and intra-class variations in LIRIS HARL dataset~\cite{liris-harl-2012}, 
we use as positive examples: the ground truth + those bounding boxes which have an overlap with the ground-truth greater then 75\%, which we think is more intuitive for complex datasets to train SVMs with more positive examples rather than only ground-truth. We achieved almost 5\% gain over SVMs classification accuracy with this training strategy.
In a similar way, we consider as negative examples only those features vectors whose associated region proposal have an        overlap smaller than 30\% with respect to the ground truth bounding boxes (possibly several) present in the frame.\\
\textbf{Training with hard negative mining.}\label{subsub:hard_mining}
We train the set of class specific linear SVMs using hard negative mining \cite{felzenszwalb-2010} to speed up the training process.
Namely, in each iteration of the SVM training step we consider only those negative features which fall within the margin of the decision boundary. 
We use the publicly available toolbox \emph{Liblinear}\footnote{\url{http://www.csie.ntu.edu.tw/~cjlin/liblinear/}.}
for SVM training and use $L2$ regularizer and $L1$ hinge-loss with the following parameter values to train the SVMs:
positive loss weight $\Constant{W_{\Variable{lp}}}=2$;
SVM regularisation constant $\Constant{C}=10^{-3}$; bias multiplier $\Constant{B}=10$. 

\subsection{Testing region proposal classifiers} \label{sec:testing}
With our actionness-ranked region proposals $\Vector{r}_{\Index{i}}$ we can extract a cropped image patch and pass it to the CNNs for feature extraction in a similar fashion as described in Sections~\ref{sec:regionProposals},~\ref{sec:cnnDescriptor}.
A prediction takes the form:
\begin{equation} \label{eg:svm_score}
s_c(\Vector{r}) =  \Vector{w}_{c}^{T}\Phi(\Vector{r}) + b_c ,
\end{equation}
where,$\Phi(\Vector{r})$ = $\{ {\Phi}_{a}(\Vector{r}); \Phi_{f}(\Vector{r})\}$ is combination of appearance and flow features of $\Vector{r}$ , $\Vector{w}_{c}^{T}$ and $b_c$ are the hyperplane parameter and the bias term of the learned SVM model of class $c$. 

The confidence measure $s_c(\Vector{r})$ that the action `$c$' has happened in region `$\Vector{r}$' is based on the appearance and flow features.
Due to the typically large number of  region proposals generated by the algorithms of Section \ref{sec:regionProposals}, we further apply non-maximum suppression to prune the regions.



\subsection{Action tube generation and classification}
\label{sec:dynamicProgramming}

Since our region proposals are generated on each video frame, linking these regions in space and time is essential to generate action tubes.
We formulate the action tube detection problem as a labelling problem which is divided into two parts:
i) we link the spatial regions into temporally connected action paths for each action, and
ii) we perform a pice-wise constant temporal labelling on the action paths.
Each region proposal $\Vector{r}_t$ at time $t$ and $\Vector{r}_{t+1}$ at time $t+1$,
is associated with a vector of scores $s_c(\Vector{r}_t)$ denoting the score of class $c \in \Set{C}$.\\

\textbf{Constructing proposal action paths.}
Linking of regions in time is first performed for each action category individually to form action paths (cf.~\ref{subsec:prob_formulation}).
We formulate the region proposal association problem into into a path finding problem,
which will produce $\Constant{K}$-connected paths for each action on the whole video, 
where $\Constant{K}$ is the minimum number of regions proposals generated in any frame of the video.
We can define an association score between those regions to be a sum of unary and pairwise potentials between adjacent regions:
\begin{equation}
E_c(\Vector{r}_{\Index{t}}, \Vector{r}_{\Index{t}+1} ) = s_c(\Vector{r}_t)
                                                        + s_c(\Vector{r}_{t+1})
                                                        +\lambda\cdot \psi(\Vector{r}_{\Index{t}}, \Vector{r}_{\Index{t}+1}),
\end{equation}
where $\psi(\Vector{r}_{\Index{t}}, \Vector{r}_{\Index{t}+1})$ is intersection-over-union of two regions $\Vector{r}_t$ and $\Vector{r}_{t+1}$and $\lambda$ is a scalar parameter weighting the relative importance of the pairwise term.
This energy value of two region proposals being linked would be high if both regions have a high score for a particular action class, and if both regions overlap significantly.
For each action class we can optimally solve for the action paths by solving:
\begin{equation}
\label{eq:region_max}
\Vector{R}_{c} = \arg\max_{\Vector{R}}  \frac{1}{T} \sum\limits_{t=1}^{T-1} E_c(\Vector{r}_{\Index{t}}, \Vector{r}_{\Index{t}+1})
\end{equation}
where $\Vector{R}_{c} = [\Vector{r}_{1},\Vector{r}_{2},......,\Vector{r}_{T}]$ is sequence of linked regions for action class $c$. We solve the energy maximisation problem (\ref{eq:region_max}) via dynamic programming.

Once the optimal path has been found, we remove all the region proposals that form the path and again find another action path until no more paths can be found.
For computational efficiency in the subsequent processing steps, we stop extract paths after finding the first three which have maximum energy.

As a result we have multiple paths for each action class in a video.
However, human action instances occupy only a fraction of time within the video.
Furthermore, instances of the same action class can take place at the same time,
and two or more actions instances from different categories may happen concurrently. 
Therefore, the temporal trimming of the proposed action paths produced by the above procedure is required to achieve action instance detection.

\paragraph{Temporal localisation.}
Although action paths are associated with individual action classes,
because of the way they are constructed (cf. Equation~\ref{eq:region_max} again) the scores of single frame-level region proposals within a path might not be consistent.
Therefore, we formulate the temporal trimming of action paths as a labelling problem \cite{Evangel-2014}.  

The goal is to assign to every region $\Vector{r}_t \in \Vector{R}_{c}$ in an action path $\Vector{R}_{c}$ a label \mbox{${c}_{t}$ $\in$ $\Constant{C}$} subject to the constraints that:
i) the sequence of labels ${c} = [c_1, c_2, \dots,c_T]$ should be consistent with the observations, 
ii) the sequence of labels ${c}$ is smooth in order to avoid sudden jumps in labels assigned to consecutive frames.
The problem can be cast into an energy maximisation framework with energy given by:
\begin{equation} \label{eq:energy-temporal}
E(c) = E_D(c) - E_S(c)
\end{equation}
where $E_D(c)$ (the data term) measures the similarity between $c$ and the observations and $E_S(c)$ (the smoothness term) penalises labellings that are not piece-wise constant (label jumps).
Under a first-order Markovian assumption, the term can be written as a summation of pairwise potentials, namely:
\begin{equation} \label{eq:smoothness_term}
E_S(c)=\sum_{t=1}^{T-1} V(c_t,c_{t+1}).
\end{equation}
The piece-wise constant labelling constraint is enforced by the following potential function:
\begin{equation} \label{eq:potential}
V(c_t,c_{t+1}) = \begin{cases} 0 &\mbox{if } c_t=c_{t+1} \\
\alpha & \mbox{otherwise} , \end{cases}
\end{equation}

where, $\alpha$ is a constant term and we set the value of $\alpha = 3$ from cross validation on the training set of LIRIS HARL dataset.
In order to efficiently solve the global optimisation problem we use a dynamic programming approach. We can fill the dynamic programming matrix $M$ of size  $|\Constant{C}|\times (T+1)$ recursively as:

\begin{equation}
M_t(c) = s_{c_t}(\Vector{r}_{t}) + \max_{c_{t-1}} M_{t-1}(c_{t-1}) - V(c_{t-1},c_t),
\end{equation}
where $c_{t-1} \in \{1,2, \dots ,C\}$.

To obtain the optimal labelling path we can back track from the maxima of the last column of $M$.
This will give us a pice-wise constant label for each region in the action path.
The labelling of this path can be written as $L_c~=~[c_1,c_2,c_3, \dots c_T]$.

\paragraph{Labelling action instances.}
As the action paths $\Vector{R}_{c}$ are associated with a specific action class $c$,
we now extract contiguous region proposal subsets which have been labelled with the same category $c$ as the action path.
The resulting subsets of action paths form the action tubes with which we perform action detection.
The set or region proposals in time (associated with $L_c$) may contain two or more segments associated with the path's action label $c$,
which we then consider as distinct action tube instances of class $c$.

For each action tube so extracted from $\Vector{R}_{c}$,
with initial frame $t_{s}$ and final frame $t_{e}$,
we have a vector of SVM scores $\Vector{S} = [ s_c(\Vector{r}_{t_s}),s_c(\Vector{r}_{t_{s+1}}), \dots ,s_c(\Vector{r}_{t_e}) ]$ for class $c$.
As a global score $\hat{s}$ for the action tube instance we take the mean of the top 10 scores in $\Vector{S}$.
We found that the mixture of average and max pooling produce robust tube scores. 
Action tubes for which the resulting global score $\hat{s}$ is less than zero are discarded, 
since a negative score produced by an SVM trained in a 1-vs-all manner indicates that the instance belongs to another category.
Also, we discard those tubes: a) which have duration $< \delta$ and b) which have average area $< \gamma$. 
Where, $\delta$ is the threshold for minimum number of frames a tube should contain to qualify for a valid detection tube.
Average area of a tube is the area computed from the average-width and -height of the boxes present in that tube, and $\gamma = \gamma_{c}/\tau$. Where $\gamma_{c}$ is the class specific average area computed from the training set, $\tau$ is the area threshold.
We set the threshold values $\delta = 20$ and $\tau=2.2$ from cross validation on training set.

%% file: text/datasets_performance_measures.tex
\section{Datasets and performance measures} \label{datasets:sec:detection}
In order to evaluate our multi-class human activity detection algorithm, 
we selected the challenging \emph{LIRIS HARL D2} human activities dataset~\cite{liris-harl-2012}.
The dataset was created for an action detection competition in which 70 teams registered.
The large number of action classes for detection compared to previous datasets\cite{laptev-2007,klaser-2010} and its difficulty meant that only two teams \cite{liris-teamp-13, liris-teamp-51} submitted results, to which we compare our results (\S~\ref{sec:results}).
The LIRIS dataset is complex because it contains image sequences containing multiple actions annotated in space and time, 
some of which occur simultaneously.
Moreover, it contains scenes where relevant human actions take place amidst other irrelevant human motion (i.e., other people performing irrelevant actions). 
The LIRIS dataset contains 10 action categories, which include human-human interactions and human-object interactions, 
for example, `discussion of two or several people', and `a person types on a keyboard'. 
A full list of categories may be found on the dataset's website\footnote{http://liris.cnrs.fr/voir/activities-dataset}.
In particular, 
we used the D2 sequences shot with a Sony camcorder with a resolution of $720 \times 576$, 
and captured at $25$ frames per second.

\subsection{Performance indicators} \label{sec:performanceIndicators} 
The qualitative and quantitative performance of our approach was computed using the evaluation tool provided for the LIRIS-HARL competition \cite{liris-harl-2012}. 
Firstly, any detected action tube  is assigned to the closest ground truth tube, based on a normalised measure of overlap over all its frames. Secondly, a detected action tube is accepted as positive if detected and ground truth tubes have the same class, and:
i) there is sufficient overlap with respect to thresholds on `spatial pixel-wise recall' $t_{sr}$, 
and `temporal frame-wise recall' $t_{tr}$, and
ii) the excess duration is sufficiently small with respect to thresholds for `spatial pixel-wise precision' $t_{sp}$, and `temporal frame-wise precision' $t_{tp}$. 

Once the four thresholds $t_{sr},t_{tr},t_{sp}$ and $t_{tp}$ are fixed, recall and precision may be calculated in the usual way as:
$\text{Recall} = \frac{\# \textnormal{correctly found actions}}{\# \textnormal{actions in ground truth}}$, 
$\text{Precision} = \frac{\# \textnormal{correctly found actions} }{ \# \textnormal{number of found actions}}$.
The F1-score combines them as:
$\Scalar{F1}~=~\frac{2\times \text{Recall} \times \text{Precision}}{\text{Recall} + \text{Precision}}$.
A final performance measure may be obtained by integrating the F1-score over the range of possible threshold values~\cite{liris-harl-2014}. 
Four integrated F1-score values ($I_{sr}, I_{sp}, I_{tr}, I_{tp}$) are first calculated by varying one threshold while setting the others to a small value ($\eta=0.1$).  Then, an overall score is obtained by averaging the four values:
\begin{equation} \label{eq:integrated-performance}
\text{Integrated Performance} = \frac{I_{sr} + I_{sp} + I_{tr} + I_{tp}}{4}
\end{equation}
which is independent from arbitrary thresholds on spatial or temporal overlap~\cite{liris-harl-2014}. 

%% file: text/experiments_results_discussion.tex
\section{Experiments results and discussion} \label{subsec:discussion}

\subsection{Results}\label{sec:results}
We evaluate two region proposal methods with our pipeline, one based on human motion segmentation (HMS) (\S~\ref{subsec:human_mot_seg}) and another one based on selective search (SS).
We will use HMS and SS abbreviations in tables and plot to show the performance of our pipeline based on each region proposal technique.
Our results are also compared to the current state-of-the-art: VPULABUAM-13~\cite{liris-teamp-13} and IACAS-51~\cite{liris-teamp-51}.

\paragraph{Instance classification performance - no localisation (NL).}
This evaluation strategy ignores the localisation information (i.e. the bounding boxes) and only focuses on whether an action is present in a video or not.
If a video contains multiple actions then system should return the labels of all the actions present correctly.
Even though our action detection framework is not specifically designed for this task, we still outperform the competition, as shown in Table~\ref{table:perf_baisc}.

\begin{figure}
  \centering
  \includegraphics[width=0.98\columnwidth]{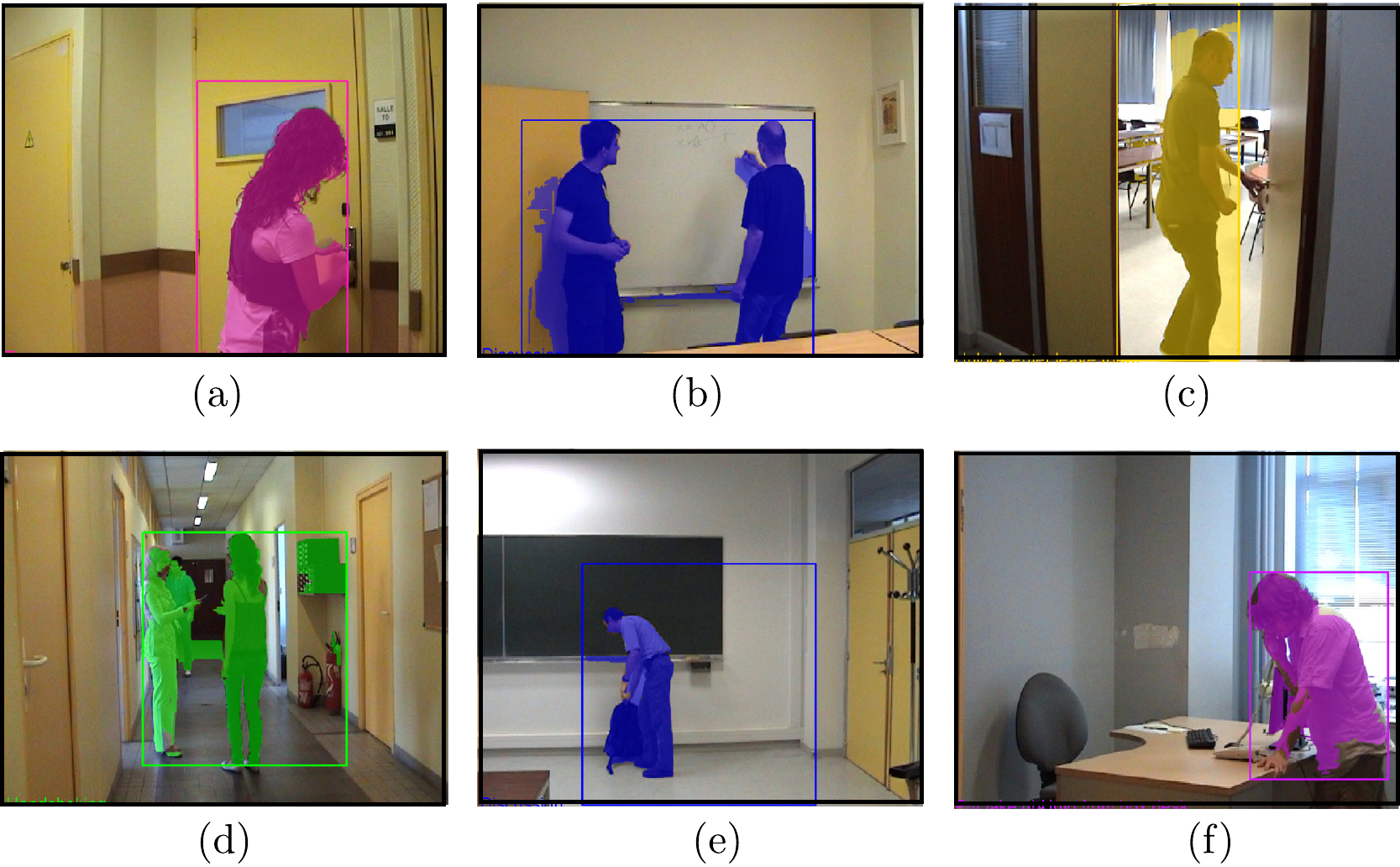}
  \caption{
    Correct (a-c) and incorrect (d-f) instance segmentation results on the LIRIS-HARL dataset, the correct category is shown in brackets.
    \textbf{(a)} `Try enter room unsuccessfully'.
    \textbf{(b)} `Discussion'.
    \textbf{(c)} `Unlock enter/leave room'.
    \textbf{(d)} `Handshaking' (Give take object from person).
    \textbf{(e)} `Discussion' (Leave bag unattended).
    \textbf{(f)} `Put take object into/from desk' (Telephone conversation).
  }
  \label{fig:actionInstanceSegmentation}
\end{figure}

\paragraph{Detection and localisation performance.}
This evaluation strategy takes localisation (space and time) information into account \cite{liris-harl-2014}.
We use a 10\% threshold quality level for the four thresholds (\S~\ref{sec:performanceIndicators}),
which is the same as that used in the LIRIS-HARL competition.
In Table~\ref{table:perf_baisc}, we denote these results as ``method-name-NL'' (NL for no localisation) and ``method-name-10\%''. 
In both cases (without localisation and with 10\% overlap),
our method outperforms existing approaches,
achieving an improvement from 46\%~\cite{liris-teamp-13} to 56\%,
in terms of F1 score without localisation measures, and a improvement from 5\%~\cite{liris-teamp-13} to 56\% (11.2 times better)
gain in the F1-score when 10\% localisation information \emph{is} taken into account.
In Table~\ref{table:perf_integrated} we list the results we obtained using the overall integrated performance scores (Equation~\ref{eq:integrated-performance}) - our method yields significantly better quantitative and qualitative results with an improvement from 3\%~\cite{liris-teamp-13} to 43\% (14.3\% times better) in terms of F1 score,
a relative gain across the spectrum of measures.
Samples of qualitative instance segmentation results are shown in Fig.~\ref{fig:actionInstanceSegmentation}.

\begin{table}
\begin{center}
\begin{tabular}{|l|c|c|c|}
\hline
Method & Recall & Precision & F1-Score \\
\hline\hline
VPULABUAM-13-NL & 0.36 & \textbf{0.66} & 0.46 \\
IACAS-51-NL & 0.3 & 0.46 & 0.36 \\
SS-NL (this work) & \textbf{0.5} &  0.53 & \textbf{ 0.52} \\
HMS-NL(this work) & \textbf{0.5} & 0.63 & \textbf{ 0.56} \\
\hline
\hline
VPULABUAM-13-10\% & 0.04 & 0.08 & 0.05 \\
IACAS-51-NL-10\% & 0.03 & 0.04 & 0.03 \\
SS-10\% (this work) & \textbf{0.5} &  0.53 & 0.52 \\
HMS-10\%(this work) & \textbf{0.5} & \textbf{0.63} & \textbf{ 0.56} \\
\hline
\end{tabular}
\end{center}
\caption{Quantitative measures precision and recall.}
\label{table:perf_baisc}
\end{table}

\begin{table}
\begin{center}
\begin{tabular}{|l|c|c|c|c|c|}
\hline
Method & $I_{sr}$ & $I_{sp}$ & $I_{tr}$ &  $I_{tp}$ & IQ \\
\hline\hline
VPULABUAM-13-IQ & 0.02 & 0.03 & 0.03 & 0.03 & 0.03\\
IACAS-51-IQ & 0.01 & 0.01 & 0.03 & 00.0 & 0.02\\
SS-IQ & 0.52 & 0.22 & 0.41 & 0.39 & 0.38 \\
HMS-IQ & 0.49 & 0.35 & 0.46 & 0.43 & \textbf{0.44} \\
\hline
\end{tabular}
\end{center}
\caption{Qualitative thresholds and integrated score. }
\label{table:perf_integrated}
\end{table}

\begin{figure*}[t]
  \centering
  \includegraphics[width=0.99\textwidth]{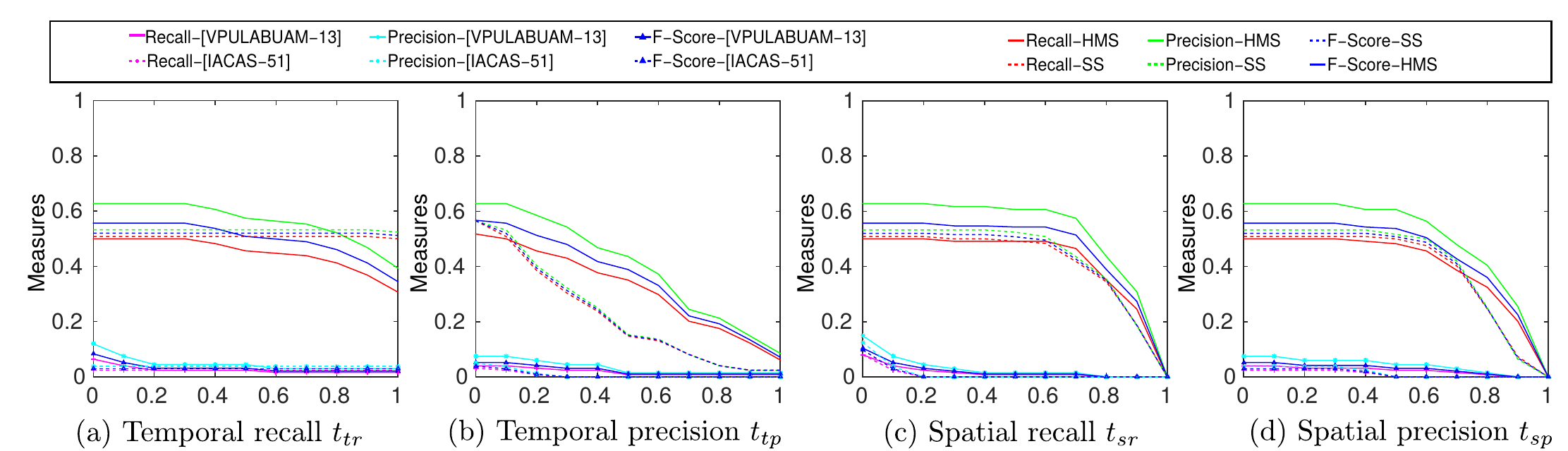}
  \caption{Performance vs detection quality curves}
  \label{fig:img_1}
\end{figure*}

\begin{figure}[t]
  \centering
  \includegraphics[width=0.99\columnwidth]{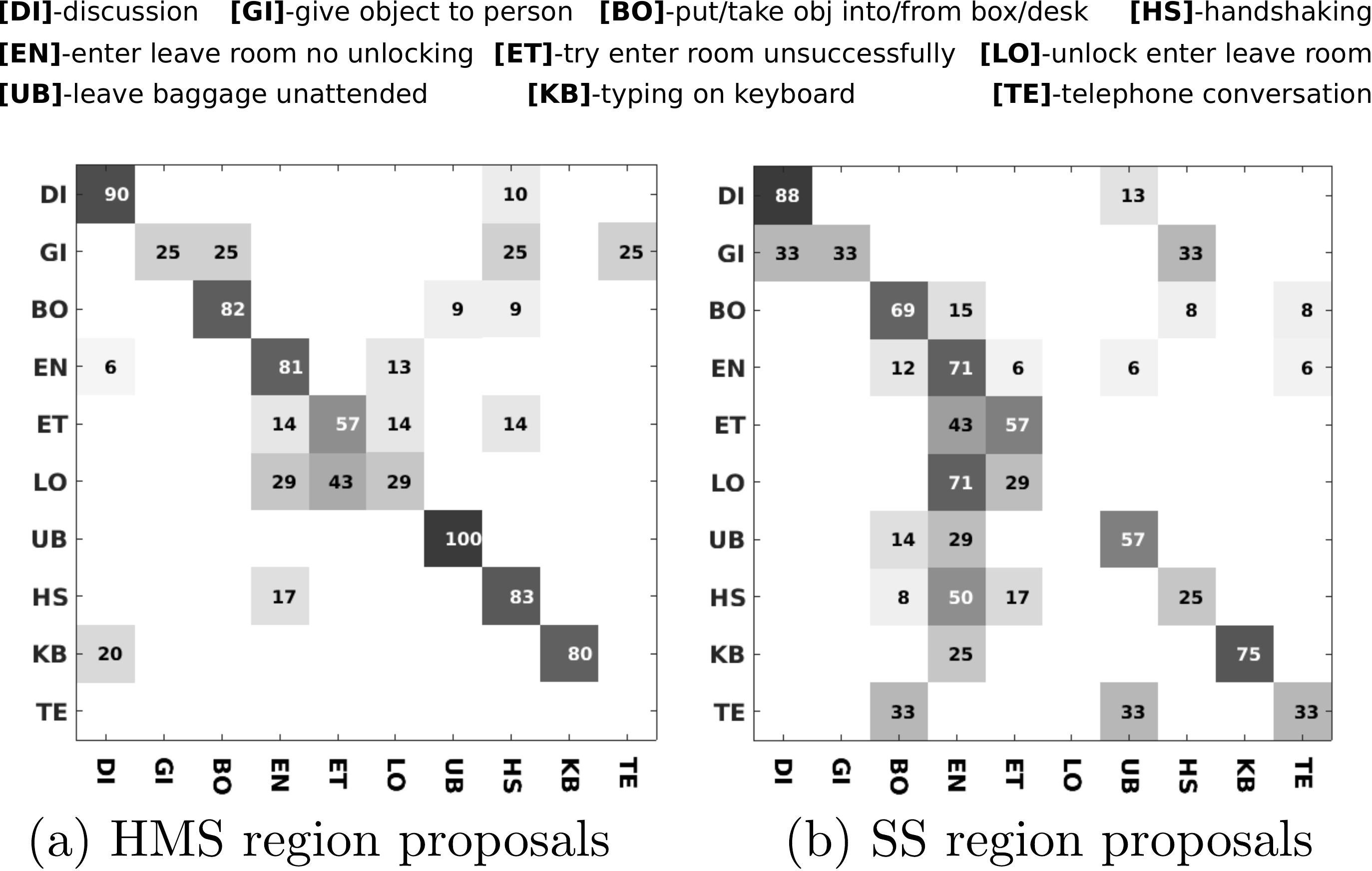}
  \caption{Confusion matrix obtained by human motion segmentation(HMS) and selective search(SS) region proposal approach. They show the classification accuracy of HMS- and SS-based methods on LIRIS HARL human activity dataset. HMS region proposal based method provides better classification accuracy on the the complex LIRIS dataset.}
  \label{fig:img_confmatrix}
\end{figure}

The pure classification accuracy of the HMS- and SS-based approaches are reflected in the Confusion Matrices shown in Figure~\ref{fig:img_confmatrix}. Confusion matrices show the the complexity of dataset. Some of the actions are completely confusing with others, for eg. telephone-conversation is completely confused with put-take-object-to-from-box-desk, same can be observed for action unlock-enter-leave-room in SS approach. 

\paragraph{Performance vs detection quality curves.}
The plots in Figure~\ref{fig:img_1} attest the robustness of our method,
as they depict the curves corresponding to precision, recall and F1-score over varying quality thresholds.

When the threshold $t_{tr}$ for temporal recall is considered (\S~Figure~\ref{fig:img_1}~plot-(a)) we achieved a highest recall
of 50\% for both HMS- and SS-based approaches and a highest precision of 65\% for HMS-based approach at threshold value of $t_{tr}$=0. 
As the threshold increases towards $t_{tr}=1$,
SS-based method shows a robust performance, with highest recall=50\% and precision=52\%, HMS-based method shows promising results with an acceptable drop in precision and recall.
Note that when $t_{tr}$=1,  we assume that all frames of an activity instance need to be detected in order for the instance itself to be considered as detected. \\
As for the competing methods, IACAS-51~\cite{liris-teamp-51} yields the next competing recall of 2.4\% and a  precision of 3.7\% with a threshold value of $t_{tr}$=1. 

When acting on the value of the temporal frame-wise precision threshold $t_{tp}$
(\S~Figure~\ref{fig:img_1}~plot-(b)) we can observe that at $t_{tp}$=1,
when we assume that not a single spurious frame outside the ground-truth temporal window is allowed,
our HMS-based region proposal approach gives highest recall of 8\% and precision 10.7\%, where, as SS-based approach has significantly lower recall=2\% and precision=2.4\%, 
which is still significantly higher than the performance of the existing methods.
Indeed, at $t_{tp}$=1, VPULABUAM-13 has recall=0.8\% and precision=1\% where IACAS-51 yields both zero precision and zero recall. This results tell us that HMS-based approach performs superior in detecting temporal extent of an action and thus is suitable for action localisation in temporally untrimmed videos.
The remaining two plots-(c) and -(d) of~Figure~\ref{fig:img_1}
 illustrate the overall performance when spatial overlap is taken into account.
 Both plots show metrics approaching zero when the corresponding spatial thresholds ({pixel-wise recall} $t_{sr}$ and {pixel-wise precision} $t_{sp}$) approach 1.
 Note that it is highly unlikely for a ground truth activity to be consistently (spatially) included in the corresponding detected activity over all the consecutive frames (spatial recall), as indicated in the plot-(c). It is also rare for a detected activity to be (spatially) included in the corresponding ground-truth activity over all the frames (spatial precision) as indicated in plot-(d).\\ 
For the pixel-wise recall ( plot-(c)), our HMS based method shows consistent recall between  45\% to 50\% and precision between 59\% to 65.5\% up to a threshold value of $t_{sr}$=0.7, where as, SS-based region proposal approach gives comparable recall between 48.3\% to 50.8\%, but relative lower precision between 43.5\% to 53.2\% upto $t_{sr}$=0.7.
For the pixel-wise precision (plot-(d)), HMS and SS-based approaches give similar recall between 39\% to 50\%, where as HMS-method again outperforms in precision with 48\% to 63\% up to a threshold value of $t_{sp}$=0.7, where as SS has precision 41\% to 53\% up to a threshold value $t_{sp}$=0.7.
Finally, we draw conclusion that our HMS-based region proposal approach shows superior qualitative and quantitative detection performance on the challenging LIRIS HARL dataset. 





%% file: text/conclusion.tex
\section{Conclusions} \label{sec:conclusions}

In this paper, we presented a novel human action recognition approach which,
unlike existing state-of-the-art approaches which typically deal with single action classification and/or localisation problems on temporally trimmed videos,
addresses in a coherent framework the challenges involved in concurrent multiple human action recognition,
spatial localisation and temporal detection.

We tested our method on the challenging LIRIS-HARL D2~\cite{liris-harl-2012} dataset which contains multiple concurrent actions,
with instances of the same action class happening at the same time, and where all videos are temporally untrimmed.
Our proposed pipeline achieved 
new benchmark performance which is 14.3 times better than the previous top performer.
By adapting our method to only use region proposals from independent frames at test time without the need for costly space-time action motion segmentation, 
we may extent our tube generation and labelling algorithm to be fully incremental and online by updating the dynamic programming optimisation for every new incoming frame. Once online, we will be able to detect actions as they happen in a live video stream.
\vfill

%% file: main.bbl
\begin{thebibliography}{10}\itemsep=-1pt

\bibitem{TRECVID}
Trecvid med 13., 2013.
\newblock http://www.nist.gov/itl/iad/mig/med13.cfm.

\bibitem{Brox-2004}
T.~Brox, A.~Bruhn, N.~Papenberg, and J.~Weickert.
\newblock High accuracy optical flow estimation based on a theory for warping.
\newblock {\em Proc. European Conf. Computer Vision}, 2004.

\bibitem{Brox_2011}
T.~Brox and J.~Malik.
\newblock Large displacement optical flow: descriptor matching in variational
  motion estimation.
\newblock {\em IEEE Transactions on Pattern Analysis and Machine Intelligence},
  33(3):500--513, 2011.

\bibitem{dalal-2005}
N.~Dalal and B.~Triggs.
\newblock Histograms of oriented gradients for human detection.
\newblock In {\em {IEEE} Int. Conf. on Computer Vision and Pattern
  Recognition}, volume~1, pages 886--893 vol. 1, June 2005.

\bibitem{dalal-2006}
N.~Dalal, B.~Triggs, and C.~Schmid.
\newblock Human detection using oriented histograms of flow and appearance.
\newblock In {\em Proc. European Conf. Computer Vision}, 2006.

\bibitem{escalera2014chalearn}
S.~Escalera, X.~Bar{\'o}, J.~Gonzalez, M.~A. Bautista, M.~Madadi, M.~Reyes,
  V.~Ponce-L{\'o}pez, H.~J. Escalante, J.~Shotton, and I.~Guyon.
\newblock Chalearn looking at people challenge 2014: Dataset and results.
\newblock In {\em Computer Vision-ECCV 2014 Workshops}, pages 459--473.
  Springer, 2014.

\bibitem{Evangel-2014}
G.~Evangelidis, G.~Singh, and R.~Horaud.
\newblock Continuous gesture recognition from articulated poses.
\newblock In {\em ECCV Workshops}, 2014.

\bibitem{felzenszwalb-2010}
P.~Felzenszwalb, R.~Girshick, D.~McAllester, and D.~Ramanan.
\newblock Object detection with discriminatively trained part based models.
\newblock {\em {IEEE} Trans.\ Pattern Analysis and Machine Intelligence},
  32(9):1627--1645, 2010.

\bibitem{girshick-2014}
R.~Girshick, J.~Donahue, T.~Darrel, and J.~Malik.
\newblock Rich feature hierarchies for accurate object detection and semantic
  segmentation.
\newblock In {\em {IEEE} Int. Conf. on Computer Vision and Pattern
  Recognition}, 2014.

\bibitem{Georgia-2015b}
G.~Gkioxari, R.~B. Girshick, and J.~Malik.
\newblock Contextual action recognition with {R*CNN}.
\newblock {\em CoRR}, abs/1505.01197, 2015.

\bibitem{Georgia-2015a}
G.~Gkioxari and J.~Malik.
\newblock Finding action tubes.
\newblock In {\em {IEEE} Int. Conf. on Computer Vision and Pattern
  Recognition}, 2015.

\bibitem{gorban2015thumos}
A.~Gorban, H.~Idrees, Y.~Jiang, A.~R. Zamir, I.~Laptev, M.~Shah, and
  R.~Sukthankar.
\newblock Thumos challenge: Action recognition with a large number of classes,
  2015.

\bibitem{liris-teamp-51}
Y.~He, H.~Liu, W.~Sui, S.~Xiang, and C.~Pan.
\newblock Liris harl competition participant, 2012.
\newblock Institute of Automation, Chinese Academy of Sciences, Beijing
  \url{http://liris.cnrs.fr/harl2012/results.html}.

\bibitem{Jain-2015}
M.~Jain, J.~C.~v. Gemert, and C.~G. Snoek.
\newblock What do 15,000 object categories tell us about classifying and
  localizing actions?
\newblock {\em CVPR}, 2015.

\bibitem{Shuiwang-2013}
S.~Ji, W.~Xu, M.~Yang, and K.~Yu.
\newblock 3d convolutional neural networks for human action recognition.
\newblock {\em Pattern Analysis and Machine Intelligence, IEEE Transactions
  on}, 35(1):221--231, Jan 2013.

\bibitem{Jia-2013}
Y.~Jia, E.~Shelhamer, J.~Donahue, S.~Karayev, J.~Long, R.~B. Girshick,
  S.~Guadarrama, and T.~Darrell.
\newblock Caffe: Convolutional architecture for fast feature embedding.
\newblock {\em CoRR}, abs/1408.5093, 2014.

\bibitem{jiang2014thumos}
Y.~Jiang, J.~Liu, A.~Roshan~Zamir, G.~Toderici, I.~Laptev, M.~Shah, and
  R.~Sukthankar.
\newblock Thumos challenge: Action recognition with a large number of classes.
\newblock {\em http://crcv.ucf.edu/THUMOS14}, 2014.

\bibitem{Karpathy-2014}
A.~Karpathy, G.~Toderici, S.~Shetty, T.~Leung, R.~Sukthankar, and L.~Fei-Fei.
\newblock Large-scale video classification with convolutional neural networks.
\newblock In {\em {IEEE} Int. Conf. on Computer Vision and Pattern
  Recognition}, 2014.

\bibitem{klaser-2010}
A.~Kl{\"a}ser, M.~Marsza{\l}ek, C.~Schmid, and A.~Zisserman.
\newblock Human focused action localization in video.
\newblock In {\em International Workshop on Sign, Gesture, Activity}, 2010.

\bibitem{krizhevsky2012}
A.~Krizhevsky, I.~Sutskever, and G.~E. Hinton.
\newblock Imagenet classification with deep convolutional neural networks.
\newblock In {\em Advances in Neural Information Processing Systems}, 2012.

\bibitem{laptev-2008}
I.~Laptev, M.~Marsza{\l}ek, C.~Schmid, and B.~Rozenfeld.
\newblock Learning realistic human actions from movies.
\newblock In {\em {IEEE} Int. Conf. on Computer Vision and Pattern
  Recognition}, 2008.

\bibitem{laptev-2007}
I.~Laptev and P.~Pérez.
\newblock Retrieving actions in movies.
\newblock In {\em Proc. Int. Conf. Comp. Vis.(ICCV'07)}, pages 1--8, Rio de
  Janeiro, Brazil, October 2007.

\bibitem{Lowe-2004}
D.~Lowe.
\newblock Distinctive image features from scale-invariant keypoints.
\newblock {\em IJCV}, 2004.

\bibitem{Lu_2015_CVPR}
J.~Lu, r.~Xu, and J.~J. Corso.
\newblock Human action segmentation with hierarchical supervoxel consistency.
\newblock In {\em {IEEE} Int. Conf. on Computer Vision and Pattern
  Recognition}, June 2015.

\bibitem{liris-teamp-13}
J.~C. SanMiguel and S.~Suja.
\newblock Liris harl competition participant, 2012.
\newblock Video Processing and Understanding Lab, Universidad Autonoma of
  Madrid, Spain, \url{http://liris.cnrs.fr/harl2012/results.html}.

\bibitem{Simonyan-2014}
K.~Simonyan and A.~Zisserman.
\newblock Two-stream convolutional networks for action recognition in videos.
\newblock In {\em Advances in Neural Information Processing Systems 27}, pages
  568--576. Curran Associates, Inc., 2014.

\bibitem{soomro-2012}
K.~Soomro, A.~R. Zamir, and M.~Shah.
\newblock {UCF101}: A dataset of 101 human action classes from videos in the
  wild.
\newblock Technical report, CRCV-TR-12-01, 2012.

\bibitem{uijlings-2013}
J.~Uijlings, K.~van~de Sande, T.~Gevers, and A.~Smeulders.
\newblock Selective search for object recognition.
\newblock {\em Int. Journal of Computer Vision}, 2013.

\bibitem{wang-2011}
H.~Wang, A.~Kl{\"a}ser, C.~Schmid, and C.~Liu.
\newblock {Action Recognition by Dense Trajectories}.
\newblock In {\em {IEEE} Int. Conf. on Computer Vision and Pattern
  Recognition}, 2011.

\bibitem{wang-2013}
H.~Wang and C.~Schmid.
\newblock {Action Recognition with Improved Trajectories}.
\newblock In {\em Proc. Int. Conf. Computer Vision}, pages 3551--3558, 2013.

\bibitem{Weinzaepfel-2015}
P.~Weinzaepfel, Z.~Harchaoui, and C.~Schmid.
\newblock {Learning to track for spatio-temporal action localization}.
\newblock In {\em {IEEE} Int. Conf. on Computer Vision and Pattern
  Recognition}, June 2015.

\bibitem{liris-harl-2012}
C.~{Wolf}, J.~{Mille}, E.~{Lombardi}, O.~{Celiktutan}, M.~{Jiu},
  M.~{Baccouche}, E.~{Dellandréa}, C.-E. {Bichot}, C.~{Garcia}, and
  B.~{Sankur}.
\newblock {The LIRIS Human activities dataset and the ICPR 2012 human
  activities recognition and localization competition}.
\newblock Technical Report RR-LIRIS-2012-004, LIRIS UMR 5205 CNRS/INSA de
  Lyon/Universit\'{e} Claude Bernard Lyon 1/Universit\'{e} Lumi\`{e}re Lyon
  2/\'{E}cole Centrale de Lyon, Mar. 2012.

\bibitem{liris-harl-2014}
C.~Wolf, J.~Mille, E.~Lombardi, O.~Celiktutan, M.~Jiu, E.~Dogan, G.~Eren,
  M.~Baccouche, E.~Dellandrea, C.-E. Bichot, C.~Garcia, and B.~Sankur.
\newblock Evaluation of video activity localizations integrating quality and
  quantity measurements.
\newblock {\em In Computer Vision and Image Understanding}, 127:14--30, 2014.

\bibitem{xu2014discriminative}
Z.~Xu, Y.~Yang, and A.~G. Hauptmann.
\newblock A discriminative cnn video representation for event detection.
\newblock {\em arXiv preprint arXiv:1411.4006}, 2014.

\bibitem{yeung2015every}
S.~Yeung, O.~Russakovsky, N.~Jin, M.~Andriluka, G.~Mori, and L.~Fei-Fei.
\newblock Every moment counts: Dense detailed labeling of actions in complex
  videos.
\newblock {\em arXiv preprint arXiv:1507.05738}, 2015.

\bibitem{Zeiler2013}
M.~D. Zeiler and R.~Fergus.
\newblock Visualizing and understanding convolutional networks.
\newblock In {\em CoRR}, 2013.

\end{thebibliography}
